\documentclass[lettersize,journal]{IEEEtran}
\usepackage{amsmath,amsfonts}
\usepackage{algorithmic}
\usepackage{algorithm}
\usepackage{array}
\usepackage[caption=false,font=normalsize,labelfont=sf,textfont=sf]{subfig}
\usepackage{textcomp}
\usepackage{stfloats}
\usepackage{url}
\usepackage{verbatim}
\usepackage{graphicx}
\usepackage{cite}
\usepackage{amsmath,graphicx}
\usepackage{adjustbox}
\usepackage{url}
\usepackage{array}
\usepackage{multirow}
\usepackage{amssymb}
\usepackage{pifont}
\usepackage{pgfplots}
\usepackage{multirow}
\usepackage{booktabs}
\usepackage{makecell}
\usepackage{tikz}
\usepackage{tipa}
\usepackage{xcolor}
\usepackage{caption}
\usepackage{subcaption}
\usepackage{lipsum}
\usepackage{arydshln}
\newcommand{\cmark}{\ding{51}}
\newcommand{\xmark}{\ding{55}}
\definecolor{lightblue}{rgb}{0.93,0.95,1.0} 
\usepackage[compact]{titlesec}
\titlespacing{\subsection}{0pt}{1.5ex plus 0.5ex minus .2ex}{0.8ex plus .2ex}
\titlespacing{\subsubsection}{0pt}{1.0ex plus 0.5ex minus .2ex}{0.5ex plus .2ex}

\begin{document}

\title{Towards Inclusive Communication: A Unified Framework for Generating Spoken Language from Sign, Lip, and Audio}
\author{Jeong Hun Yeo, Hyeongseop Rha, Sungjune Park, Junil Won, and Yong Man Ro,~\IEEEmembership{Fellow,~IEEE}

\thanks{J. H. Yeo, H. Rha, S. Park, J. Won, and Y. M. Ro are with Integrated Vision Language Laboratory, School of Electrical Engineering, Korea Advanced Institue of Science and Technology (KAIST), Republic of Korea (e-mail: sedne246@kaist.ac.kr; ryool\_1832@kaist.ac.kr; sungjune-p@kaist.ac.kr; dnjswnsdlf48@kaist.ac.kr; ymro@kaist.ac.kr), Corresponding author: Y. M. Ro (fax: 82-42-350-5494)}
}




\maketitle

\begin{abstract}
Audio is the primary modality for human communication and has driven the success of Automatic Speech Recognition (ASR) technologies. However, such audio-centric systems inherently exclude individuals who are deaf or hard of hearing. Visual alternatives such as sign language and lip reading offer effective substitutes, and recent advances in Sign Language Translation (SLT) and Visual Speech Recognition (VSR) have improved audio-less communication. Yet, these modalities have largely been studied in isolation, and their integration within a unified framework remains underexplored. In this paper, we propose the first unified framework capable of handling diverse combinations of sign language, lip movements, and audio for spoken-language text generation. We focus on three main objectives: (i) designing a unified, modality-agnostic architecture capable of effectively processing heterogeneous inputs; (ii) exploring the underexamined synergy among modalities, particularly the role of lip movements as non-manual cues in sign language comprehension; and (iii) achieving performance on par with or superior to state-of-the-art models specialized for individual tasks. Building on this framework, we achieve performance on par with or better than task-specific state-of-the-art models across SLT, VSR, ASR, and Audio-Visual Speech Recognition. Furthermore, our analysis reveals a key linguistic insight: explicitly modeling lip movements as a distinct modality significantly improves SLT performance by capturing critical non-manual cues.
\end{abstract}

\section{Introduction}
Audio plays a central role in human communication and serves as the primary modality for conveying linguistic meaning. Accordingly, Automatic Speech Recognition (ASR) technologies have seen rapid advancement ~\cite{amodei2016deep,kim2017joint,prabhavalkar2023end}, enabling seamless voice-based interaction with smart devices in daily life.

However, audio-based interfaces remain fundamentally inaccessible to individuals who are deaf or hard of hearing. For such users, sign language and lip movements serve as effective visual alternatives. Sign language is a gesture-based linguistic system with its own grammar and syntax, while lip reading allows the interpretation of spoken language through the observation of mouth movements. Recent research has leveraged these alternatives through Sign Language Translation (SLT)~\cite{chen2024factorized, jiao2024visual, rust2024towards, jang2025deep, zhou2021spatial} and Visual Speech Recognition (VSR)~\cite{ma2022visual, kim2023lip, kim2021cromm}, making audio-less communication more accessible.

Building on this progress in each area, recent efforts have aimed at integrating multiple modalities into a unified architecture~\cite{makino2019recurrent, hsu2022u, haliassos2024unified}. While these efforts mark a step towards unification, they have been predominantly limited to the audio-lip fusion seen in Audio-Visual Speech Recognition (AVSR)~\cite{afouras2018deep, tao2020end}. This narrow focus overlooks the broader potential of combining sign language, lip movements, and audio. A unified framework that supports flexible combinations of these modalities not only enables more inclusive interaction for users with varying hearing capabilities but also overcomes the limitations of treating each modality in isolation, such as redundant training pipelines, siloed representations, and restricted cross-modal learning. Furthermore, this integration paves the way for more efficient training and richer multimodal representations that reflect the natural synergy between modalities in human communication.

In this paper, we present a unified framework that supports various input combinations of sign language, lip movements, and audio, enabling a single model to generate spoken-language text across a range of communication scenarios, particularly SLT, VSR, ASR, and AVSR. As the pioneering study to jointly investigate these modalities within a single system, we pursue three primary objectives: (i) to design a simple and unified architecture for processing the three modalities, (ii) to explore the previously underexamined synergy between these modalities, particularly by investigating the role of lip movements as non-manual linguistic cues in sign language, and (iii) to achieve performance comparable to task-specific models using a single model. To achieve these goals, we design a system that aligns and integrates heterogeneous modalities into a unified representation. Specifically, we start from the insight that input modalities differ in form but share the same objective of conveying linguistic content. Prior work in AVSR~\cite{zhu2023vatlm, yeo2025mms} has shown that although lip movements and audio differ in granularity, such as phonemes versus visemes (i.e., the visual counterparts of phonemes), their frame-level features can be temporally aligned and jointly leveraged through simple fusion strategies. Building on this observation, we hypothesize that frame-level sign features, which contain gloss information (i.e., representations of individual sign meanings), can likewise be temporally aligned and integrated. Based on this hypothesis, we model a unified linguistic representation by aligning and fusing the temporally synchronized features from the three modalities. These unified representations are then passed to a Large Language Model (LLM) decoder, leveraging recent findings that LLMs are capable of context-aware decoding and cross-modal linguistic interpretation~\cite{gong2024llms,cappellazzo2025large}.

Beyond its practical utility, this unified setting enables analytical insights into the contribution of individual modalities. In particular, it allows us to systematically examine the impact of incorporating lip movements into SLT as a distinct modality, rather than treating them as part of general visual input. Although non-manual cues in sign language, especially lip movements, are known to convey grammatical and semantic distinctions~\cite{sandler2009symbiotic}, they have been computationally underutilized in prior work, primarily due to the lack of architectural frameworks capable of treating them as a distinct input stream. Our work directly addresses this architectural gap. The proposed model reveals that leveraging lip movements as a complementary modality significantly improves translation quality, highlighting their essential role in sign language communication.

However, training a unified model across heterogeneous modalities presents new challenges. While the model performs well on audio-based tasks such as ASR and AVSR, it learns more slowly on visually grounded tasks like VSR and SLT, likely due to the higher complexity of extracting linguistic information from sign gestures and lip movements. Notably, VSR shows limited generalization when trained jointly. To mitigate this, we introduce a two-stage training strategy: we first focus on visual tasks to ensure more stable learning, and then fine-tune the model jointly across all modalities to balance performance. Beyond performance, our approach offers new possibilities for building inclusive communication tools, especially for users with varying hearing abilities, by enabling flexible interaction through multiple input types.

The contributions of this paper can be summarized as:
\begin{itemize}
    \item \textbf{A Unified Tri-Modal Framework}  
    A single architecture that flexibly integrates sign language, lip movements, and audio to perform SLT, VSR, ASR, and AVSR within one model. 

    \item \textbf{Novel Linguistic Insight on SLT}  
    The first empirical evidence that explicitly modeling lip movements as a separate modality significantly enhances sign language translation by capturing non-manual cues. 

    \item \textbf{An Effective Two-Stage Training Strategy}  
    A specialized training schedule that resolves the learning imbalance between visual and audio tasks, optimizing performance across all modalities. 
\end{itemize}

\section{Related Works}
\subsection{Automatic Speech Recognition}
As audio is a core modality for conveying linguistic meaning, Automatic Speech Recognition (ASR) is among the first tasks to benefit from deep learning, well before research into visual modalities such as sign language and lip reading. 

Advances in deep learning architectures have played a crucial role in improving ASR performance. Early studies replaced traditional HMM-GMM acoustic models with Deep Neural Networks (DNNs)~\cite{hinton2012deep}. To further capture local time-frequency patterns, Convolutional Neural Networks (CNNs)~\cite{abdel2014convolutional} were introduced, while Recurrent Neural Networks (RNNs)~\cite{graves2013speech} model sequential dependencies. More recently, attention-based Transformer~\cite{vaswani2017attention} architectures have enabled the integration of long-range context, contributing to substantial performance gains.

In addition to architectural advances, the availability of large-scale datasets~\cite {panayotov2015librispeech} and improved learning methods have also played key roles in the progress of ASR. Techniques such as Connectionist Temporal Classification (CTC)~\cite{graves2006connectionist} and sequence-to-sequence~\cite{sutskever2014sequence} learning have enabled end-to-end speech recognition. While supervised approaches require large labeled datasets, self-supervised learning methods~\cite{baevski2020wav2vec, hsu2021hubert} have been developed to leverage vast amounts of unlabeled audio. In parallel, Whisper~\cite{radford2023robust}, trained on 680,000 hours of multilingual and noisy speech via large-scale supervised learning, has emerged as a powerful foundation model for speech recognition. With the rise of powerful audio encoders, researchers~\cite{chu2023qwen} have begun integrating LLMs~\cite{achiam2023gpt, touvron2023llama} into ASR pipelines, enhancing contextual understanding and recognition quality.

\subsection{Visual Speech Recognition}
Despite the importance of lip movements for individuals who are deaf or hard of hearing, Visual Speech Recognition (VSR) received comparatively less attention than ASR in its early stages, primarily due to the difficulty of collecting and annotating large-scale video data. Recent releases of large-scale datasets~\cite{afouras2018lrs3, chung2016lip, sheng2022importance, sheng2021adaptive} have accelerated significant progress in VSR research. 

VSR architectures have followed a development path broadly similar to ASR, evolving from CNNs~\cite{noda2014lipreading} for spatiotemporal feature extraction, to RNNs~\cite{petridis2017end} for temporal modeling, and more recently to Transformer-based models~\cite{afouras2018deep} that capture long-range dependencies. However, VSR poses unique challenges. A notable example is the prevalence of homophenes (i.e., different phonemes that appear visually similar)~\cite{kim2022distinguishing}, which introduce ambiguity and limit performance. To overcome these challenges, multimodal approaches that incorporate audio signals as auxiliary supervision have become increasingly popular, leveraging the temporal alignment and rich acoustic cues of audio. These include knowledge distillation from pretrained ASR teachers~\cite{ren2021learning}, and self-supervised learning using audio-visual speech units as pseudo labels~\cite{shi2022learning, haliassos2022jointly}, and pseudo-labeling approaches~\cite{ma2023auto, yeo2024visual2} that generate text transcriptions using pretrained ASR models. Building on these advances, recent work~\cite{yeo2024visual, cappellazzo2025large} has explored integrating LLMs into VSR pipelines, enabling contextual reasoning over visually ambiguous inputs and improving robustness in real-world scenarios.

\subsection{Sign Language Translation}
Sign Language Translation (SLT) builds on prior tasks like Isolated and Continuous Sign Language Recognition (ISLR, CSLR), which extract gloss representations from sign videos. These areas have advanced significantly thanks to datasets for ISLR~\cite{li2020word} and CSLR~\cite{koller2015continuous}, enabling large-scale deep learning models~\cite{adaloglou2021comprehensive,cui2019deep,ahn2024slowfast, jang2023self}. Both tasks use deep spatiotemporal architectures like I3D~\cite{carreira2017quo} and Video Swin Transformers~\cite{liu2022video}, but differ in objectives: ISLR classifies segmented clips via cross-entropy, while CSLR~\cite{chen2022two, jang2022signing} handles unsegmented input using gloss-level supervision and CTC loss~\cite{graves2006connectionist}.

With the availability of SLT datasets~\cite{duarte2021how2sign} and visual representations learned through ISLR and CSLR, SLT research has seen significant progress. While early approaches primarily relied on visual backbones pretrained on CSLR tasks, recent work has increasingly focused on gloss-free SLT methods. This shift is largely driven by the growing difficulty of obtaining manual gloss annotations as datasets scale~\cite{uthus2023youtube}. In response, two main gloss-free approaches have emerged to effectively train the visual backbone without relying on gloss annotations: contrastive learning for visual-text alignment~\cite{jiao2024visual, zhou2023gloss, ye2024improving}, and pseudo-gloss generation for weak supervision~\cite{prajwal2022weakly, guo2025bridging}. In parallel, increasing attention has been directed toward improving the decoding stage through the integration of LLMs~\cite{gong2024llms, wong2024sign2gpt}. Since gloss sequences differ grammatically from natural language, LLM-based decoders offer clear advantages in generating fluent and natural spoken-language text. Building on the strengths of LLM-based decoders, more recent work~\cite{jang2025lost} leverages contextual cues such as preceding sentences or background descriptions.

Despite this notable progress, existing methods treat the visual modality as a monolithic input stream and do not explicitly distinguish between hand gestures and lip movements. Different from this, we integrate a dedicated VSR pipeline into the SLT framework and demonstrate that modeling lip movements through a separate pathway plays a critical role in improving sign language comprehension.


\subsection{Unified Modeling across Modalities}
Unified models for performing multi-task VSR, ASR, and AVSR have been explored in several studies. One early approach was based on RNN-T~\cite{makino2019recurrent}, trained on a large-scale audio-visual dataset (31k hours) and incorporating strategies such as modality dropout during training. However, its performance significantly lagged behind that of task-specific models. AV-HuBERT~\cite{shi2022learning} was the first to achieve performance comparable to task-specific models by leveraging a single encoder trained with self-supervised learning and modality dropout. However, it still relied on task-specific decoders. Building on this encoder, u-HuBERT~\cite{hsu2022u} extended the modality dropout strategy to the fine-tuning stage, enabling the use of a single encoder-decoder architecture and achieving strong performance across all tasks. More recently, USR~\cite{haliassos2024unified} introduced a semi-supervised learning strategy that utilizes online pseudo-labels generated by an exponential moving average (EMA)-based teacher during the fine-tuning stage. This approach demonstrated its effectiveness by achieving state-of-the-art performance on LRS3.

However, these approaches have focused on leveraging both audio inputs and lip movements to improve recognition performance in ASR, VSR, and AVSR. Beyond these two modalities, sign language also plays a crucial role in enabling inclusive communication, yet it has received comparatively less attention in unified modeling efforts. In this context, we propose a unified framework that, to the best of our knowledge, is the first to explore the integration of sign language, lip movements, and audio within a single model.

\section{Method}
\begin{figure*}[t]
\centering
\centerline{\includegraphics[width=18cm]{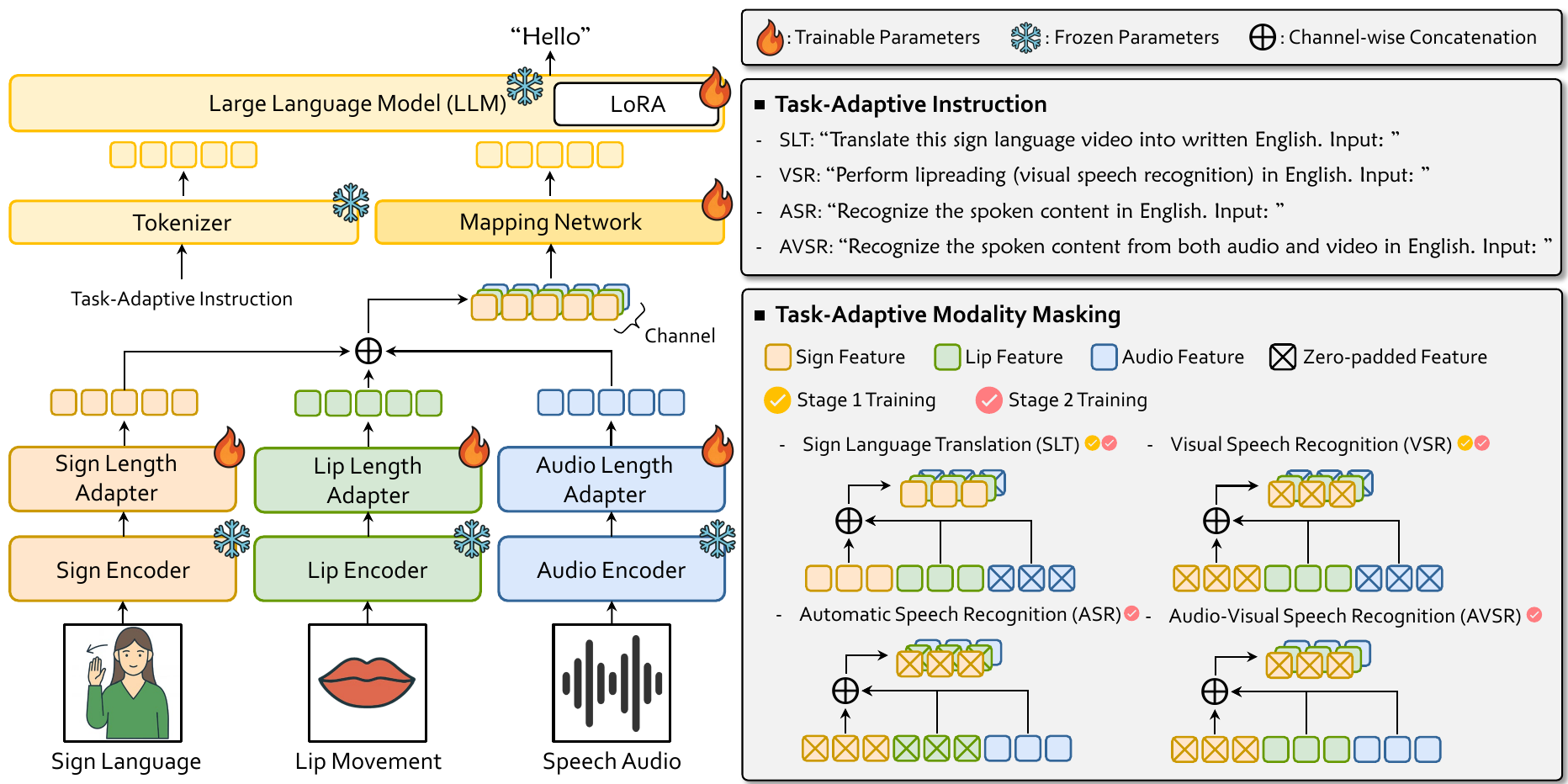}}
\caption{Overview of our proposed method. Multimodal inputs (sign, lip, audio) are temporally aligned, fused into linguistic tokens, and processed by an LLM to generate language outputs guided by task-adaptive instructions. Note that the actual input modalities vary depending on the task type.}
\label{fig:1}
\end{figure*}

In this section, we first present an overview of the proposed tri-modal framework, followed by detailed descriptions of its components. We describe the modality encoders, temporal alignment using length adapters, and the construction of unified linguistic representations via a mapping network. Finally, we introduce our multi-task training strategy across SLT, VSR, ASR, and AVSR.

\subsection{Framework Overview}
Our unified framework is capable of performing four tasks, SLT, VSR, ASR, and AVSR, by leveraging multimodal inputs: sign language, lip movements, and audio. As illustrated in Figure~\ref{fig:1}, the system consists of three modality encoders that extract frame-level features from each input stream. To address differences in temporal resolution across modalities, we employ three length adapters that transform feature sequences into a shared temporal scale. Once the features are temporally aligned, a mapping network integrates them into a unified linguistic representation, which we refer to as linguistic tokens in the remainder of this paper. These tokens are then passed to an LLM decoder, which generates spoken-language content in textual form.

\subsection{Modality Encoders}
In the following, we describe how frame-level features are extracted from each modality, capturing gloss, viseme, and phoneme information, respectively. For clarity, we refer to these features as sign, lip, and audio features throughout the paper. While the term visual encoder is commonly used in SLT and VSR tasks, we refer to our encoders as sign and lip encoders to clearly distinguish between the two visual modalities.

\subsubsection{Sign Encoder}
To generate sign features that capture gloss information, we adopt a sign encoder based on the Video Swin Transformer, following~\cite{jang2025lost}. The encoder is pre-trained on the ISLR task, which aims to predict gloss annotations from short signing video clips. Given this pre-trained sign encoder and input sign videos  $X_s = \{x_1, ..., x_{T_s}\}$, where $T_s$ denotes the number of video frames, the sign encoder $E_s$ extracts a sequence of sign features by capturing spatiotemporal patterns from the signing video. The resulting feature sequence is denoted as $F_s = E_s(X_s)$, where $F_s \in \mathbb{R}^{T_s \times D_s}$ represents sign features with the same temporal resolution as the input.

\subsubsection{Lip Encoder}
Following~\cite{yeo2024visual}, which demonstrates that lip features generated by self-supervised models can be effectively aligned with LLMs, we employ AV-HuBERT~\cite{shi2022learning} as our lip encoder. Given an input lip video $X_v= \{x_1, ..., x_{T_v}\}$, where $T_v$ denotes the number of frames, the lip encoder $E_v$ extracts a sequence of lip features that capture viseme information by modeling the temporal dynamics of lip movements. The resulting feature sequence is denoted as $F_v = E_v(X_v)$, where $F_v \in \mathbb{R}^{T_v \times D_v}$ represents lip features with the same temporal resolution as the input.

\subsubsection{Audio Encoder}
Similarly, motivated by recent studies~\cite{chu2023qwen} showing that audio features extracted by Whisper~\cite{radford2023robust} can also be effectively aligned with LLMs, we adopt it as our audio encoder. Given a log-Mel spectrogram input  \( X_a = \{x_1, ..., x_{T_a'}\} \), where $T_a'$ denotes the number of audio frames and each $x_t \in \mathbb{R}^B$ is a feature vector of $B$ Mel-frequency bins, the audio encoder $E_a$ extracts a sequence of audio features that encode phoneme information. The resulting feature sequence is denoted as $F_a = E_a(X_a)$, where $F_a \in \mathbb{R}^{T_a \times D_a}$ represents audio features. Note that $T_a'$ differs from $T_a$ due to internal downsampling applied within the Whisper encoder.

\subsection{Modality Alignment \& Fusion}
\subsubsection{Length Adapters}
Since the frame-level features capturing gloss, viseme, and phoneme information from each input modality have different temporal resolutions, we employ modality-specific length adapters to temporally align the feature sequences. These adapters are implemented using 1D convolution layers with stride values selected according to the original frame rates of the respective modalities, as detailed in the Appendix.

We denote the sign, lip, and audio length adapters as \( \text{LA}_s \), \( \text{LA}_v \), and \( \text{LA}_a \), respectively.  
Given the frame-level features \( F_s \in \mathbb{R}^{T_s \times D_s} \), \( F_v \in \mathbb{R}^{T_v \times D_v} \), and \( F_a \in \mathbb{R}^{T_a \times D_a} \),  
length adapters transform them into a unified temporal resolution \( T \) as follows:
\[
\tilde{F}_s = \text{LA}_s(F_s), \quad
\tilde{F}_v = \text{LA}_v(F_v), \quad
\tilde{F}_a = \text{LA}_a(F_a)
\]
where \( \tilde{F}_m \in \mathbb{R}^{T \times D_m} \) for each modality \( m \in \{s, v, a\} \). Here, each adapter $\text{LA}_m$ aligns the temporal resolution of the input features to the shared length $T$, while preserving the original feature dimensionality $D_m$ of each modality.

\subsubsection{Concatenation \& Mapping to Unified Linguistic Tokens}
After temporal alignment, we concatenate all modality features along the channel dimension to preserve their respective information while forming a joint representation:
\[
    \tilde{F}_{\text{concat}} = [\tilde{F}_s, \tilde{F}_v, \tilde{F}_a] \in \mathbb{R}^{T \times (D_s + D_v + D_a)}
\]
Then, we project the concatenated features into the LLM's embedding space using a shared mapping network \( \text{Map}(\cdot) \), implemented as a multi-layer perceptron (MLP):
\[
U = \text{Map}(\tilde{F}_{\text{concat}}) \in \mathbb{R}^{T \times D_{\text{LLM}}}
\]
The resulting projected features, termed unified linguistic tokens, serve as linguistic representation input compatible with LLMs.

\subsection{Multi‑Task Training Strategy} 
\subsubsection{Task‑Adaptive Modality Masking}
Since the input modalities vary across tasks, we first describe how inputs are constructed in a task-specific manner. To ensure a unified input structure across different task types, we employ a modality masking strategy by applying zero-padding to the unused modality features. For SLT, we use both sign and lip features, while the audio modality is zero-padded. For VSR, only lip features are used, with sign and audio modalities zero-padded. ASR relies solely on audio features, with sign and lip modalities zero-padded. AVSR combines lip and audio features, while the sign modality is zero-padded. These task-specific features are then projected into unified linguistic tokens using a shared mapping network as follows:
\[
\textbf{SLT}: U^{\text{SLT}} = \text{Map}([\tilde{F}_s,\ \tilde{F}_v,\ \mathbf{0}])
\]
\[
\textbf{VSR}: U^{\text{VSR}} = \text{Map}([\mathbf{0},\ \tilde{F}_v,\ \mathbf{0}])
\]
\[
\textbf{ASR}: U^{\text{ASR}} = \text{Map}([\mathbf{0},\ \mathbf{0},\ \tilde{F}_a])
\]
\[
\textbf{AVSR}: U^{\text{AVSR}} = \text{Map}([\mathbf{0},\ \tilde{F}_v,\ \tilde{F}_a])
\]
\(\mathbf{0}\) denotes a zero tensor with the same temporal resolution as the input features and the corresponding modality-specific channel dimension. Given these unified tokens, the LLM is trained in a multi-task manner with task-adaptive instructions (e.g., prompt-based task identifiers) as illustrated in Figure~\ref{fig:1}, to generate spoken language transcriptions.

\subsubsection{Two‑Stage Multi-Task Training}
When the proposed multi-task learning framework was applied to all tasks, we observed that visual-based tasks (VSR and SLT) required more training resources than audio-based tasks to achieve optimal performance. To address this imbalance, we introduce a two-stage training strategy. In the first stage, the model is trained exclusively on VSR and SLT, helping the LLM better capture the linguistic cues from sign language and lip movements. In the second stage, we perform joint training across all four tasks. Concurrently, we find that maintaining an appropriate task sampling ratio, particularly ensuring sufficient exposure to VSR, is crucial to preserving the gains from the first stage. We therefore empirically adjust these ratios to balance learning across tasks.

\section{Experimental Setup}
\subsection{Dataset}
\noindent\textbf{How2Sign}~\cite{duarte2021how2sign} is a widely used American Sign Language (ASL) dataset for the SLT task. It comprises 80 hours of instructional videos covering 10 different topics. The manually re-aligned video clips are used for training and evaluation, consisting of 31,128 sentences for training, 1,741 for validation, and 2,322 for testing.

\noindent\textbf{LRS3}~\cite{afouras2018lrs3} is a widely used English dataset for ASR, VSR, and AVSR. It comprises approximately 433 hours of TED Talk videos with aligned audio and human-annotated transcriptions. The dataset contains 151K utterances for training and 1,321 utterances for testing.

\subsection{Evaluation Metrics}
To evaluate VSR, ASR, and AVSR performance, we employ Word Error Rate (WER), a widely used metric in prior research~\cite{afouras2018deep, ma2023auto}. A lower WER reflects better recognition accuracy. For SLT evaluation, inspired by prior studies~\cite{rust2024towards}, we utilize BLEU-4~\cite{papineni2002bleu}, ROUGE~\cite{lin2004rouge}, and BLEURT~\cite{sellam2020bleurt}, where higher scores correspond to better translation quality.

\subsection{Implementation details}
Videos are resampled to 25 FPS and audio signals to 16 kHz. For sign language videos, signer regions are cropped following~\cite{jang2025lost}, and lip regions are extracted using a RetinaFace-based facial landmark detector~\cite{deng2020retinaface}, producing $96{\times}96$ patches. For the LRS3 dataset, we apply the same preprocessing to lip videos and extract log-Mel spectrograms from the audio.The sign encoder is based on Video Swin-T~\cite{liu2022video}, fine-tuned on the How2Sign dataset~\cite{jang2025lost}, producing 25 frame-level features of dimension 768. The lip encoder uses AV-HuBERT Large~\cite{shi2022learning}, pre-trained on LRS3 and VoxCeleb2~\cite{chung2018voxceleb2}, outputting 25 feature vectors per second of dimension 1024. The audio encoder is Whisper-Medium~\cite{radford2023robust}, producing 50 features per second of dimension 1024. To unify temporal resolution, we employ modality-specific 1D convolutions to downsample all features to 12.5 Hz: kernel size/stride 4 for audio and 2 for lip and sign features. The resulting features (2816-dim) are concatenated and passed through a two-layer linear projection (intermediate dimension 2944, output 3072) to match the embedding size of LLaMA 3.2-3B~\cite{touvron2023llama}. We use LLaMA 3.2-3B as the decoder, fine-tuned with LoRA~\cite{hu2021lora} (rank 16, scaling factor 32) applied to the query, key, value, and output projection layers.

\subsection{Training and Evaluation}
We apply data augmentation to the VSR and AVSR tasks by randomly cropping mouth patches to $88{\times}88$ and applying horizontal flipping. For the SLT task, we adopt text-level augmentation by randomly dropping 0–20\% of the words in the ground-truth target sentences, following~\cite{jang2025lost}. During Stage 2 training, we further apply babble noise from the NOISEX dataset~\cite{varga1993assessment} to the audio input with a probability of 75\% for the ASR and AVSR tasks, where the signal-to-noise ratio (SNR) is uniformly sampled from $\{-5, 0, 5, 10, 15, 20\}$ dB. Training proceeds in two stages: \textbf{Stage 1} trains the model on VSR and SLT tasks using a tri-stage learning rate scheduler with 18K warm-up steps, 36K decay steps, and a peak learning rate of 1e-4. \textbf{Stage 2} jointly trains ASR, VSR, AVSR, and SLT tasks with the same scheduler structure but 500 warm-up steps and 29.5K decay steps. All experiments are conducted on 8 NVIDIA A6000 GPUs (48GB each), with a maximum of 1,250 video frames processed per GPU per batch. A fixed random seed of \textbf{1} is used for reproducibility. We save two checkpoints during training: (i) the \textit{best} checkpoint based on next-token prediction accuracy on the validation set, and (ii) the \textit{last} checkpoint at the end of training. The final model is selected by comparing both checkpoints based on BLEU-4 (for SLT) and WER (for ASR, VSR, and AVSR). Due to computational constraints, we report results from a single run. For evaluation, we use beam search (width 5, temperature 0.3) for ASR, VSR, and AVSR tasks, and greedy decoding for SLT.

\section{Experimental Results}

\subsection{Comparison with the state-of-the-art methods}
While our primary objective is to develop a unified model for SLT, VSR, ASR, and AVSR, it is equally important to ensure that it achieves competitive performance compared to task‑specific models. To this end, we evaluate the performance of our unified framework on each task.

In Table~\ref{tab:1}, we compare the SLT performance of our unified framework against prior methods on the How2Sign dataset. Compared to PGG-SLT~\cite{guo2025bridging}, the previous best model without using external data, our approach improves BLEU-4 by +1.5 (15.2 vs. 13.7) and ROUGE by +3.3 (36.2 vs. 32.9). It also outperforms LiTFiC~\cite{jang2025lost} by +1.8 BLEURT (47.1 vs. 45.3). These results highlight the advantage of jointly modeling SLT and VSR within a single framework.

In Table~\ref{tab:2}, we compare performance on VSR, ASR, and AVSR using the LRS3 dataset. Since our model is LLM-based, we focus on comparisons with recent LLM-based approaches, all of which are task-specific. For VSR, our method achieves a WER of 25.9\%, comparable to task-specific models such as VSP-LLM and Llama-AVSR. In ASR, our model reaches a WER of 0.79\% using only 433 hours of training data, matching the performance of Llama-AVSR, which was trained on 1,759 hours. For AVSR, our model obtains a WER of 0.93\%, close to MMS-Llama (0.90\%). Compared to USR~\cite{haliassos2024unified}, a recent unified model for VSR, ASR, and AVSR, our method performs better on ASR (0.79\% vs. 1.2\%) and AVSR (0.93\% vs. 1.1\%) while using significantly less training data. These results demonstrate that a single unified model can effectively process sign language, lip movements, and audio signals, achieving competitive or superior performance compared to task-specific models across all four tasks. Moreover, the strong performance across tasks suggests that shared linguistic representations and cross-modal learning can generalize effectively across modalities.

\begin{table}[t]
\renewcommand{\arraystretch}{1.3}
\renewcommand{\tabcolsep}{3mm}
\centering
\caption{Performance comparison of SLT models on the How2Sign.  
$\dagger$ indicates models that are pretrained on a larger ASL corpus, YouTube-ASL~\cite{uthus2023youtube}, which contains 984 hours of video. “AV Input” indicates whether the model handles audio or lip movement modalities.}
\resizebox{0.999\linewidth}{!}{
\begin{tabular}{ccccc}
    \toprule
    \textbf{Method} 
    & \textbf{AV Input} 
    & \textbf{BLEU-4} $\uparrow$
    & \textbf{ROUGE} $\uparrow$
    & \textbf{BLEURT} $\uparrow$ \\
    \midrule
     SSLT$^\dagger$~\cite{zhang2024scaling} & \xmark & - & - & 55.7 \\
     Youtube-ASL$^\dagger$~\cite{uthus2023youtube} & \xmark & 12.4 & - & - \\
     Uni-Sign$^\dagger$~\cite{li2025uni} & \xmark & 14.9 & 36.0 & 49.4 \\
     SSVP-SLT$^\dagger$~\cite{rust2024towards} & \xmark & 15.5 & 38.4 & 49.6 \\
    \hdashline
    SSLT~\cite{zhang2024scaling} & \xmark & - & - & 34.0 \\
    SSVP-SLT~\cite{rust2024towards} & \xmark & 7.0  & 25.7 & 39.3 \\
    Fla-LLM~\cite{chen2024factorized} & \xmark & 9.7  & 27.8 & - \\
    VAP~\cite{jiao2024visual} & \xmark & 12.9  & 27.8 & - \\
    LiTFiC~\cite{jang2025lost} & \xmark & 12.7  & 32.5 & 45.3 \\
    PGG-SLT~\cite{guo2025bridging} & \xmark & 13.7  & 32.9 & - \\
    \hdashline
    Ours & \cmark & 15.2 & 36.2 & 47.1 \\
    \bottomrule
\end{tabular}}
\label{tab:1}
\end{table}

\begin{table}[t]
\renewcommand{\arraystretch}{1.3}
\renewcommand{\tabcolsep}{1.5mm}
\centering
\caption{Performance comparison of VSR, ASR, and AVSR models on LRS3. $\dagger$ indicates models that leverage pseudo-labels generated from the VoxCeleb2~\cite{chung2018voxceleb2} dataset (1,326 hours) during fine-tuning. SL denotes sign language.}
\resizebox{0.999\linewidth}{!}{
\begin{tabular}{ccccccc}
    \toprule
    \multirow{2.5}{*}{\textbf{Method}}
    & \multirow{2.5}{*}{\makecell{\textbf{SL Input} \\ \textbf{Supported}}}
    & \multirow{2.5}{*}{\makecell{\textbf{Single} \\ \textbf{Model}}}
    & \multirow{2.5}{*}{\makecell{\textbf{Training} \\ \textbf{Data (h)}}}
    & \multicolumn{3}{c}{\textbf{WER(\%)} $\downarrow$} \\
    \cmidrule(lr){5-7}
    & & & & \textbf{VSR} & \textbf{ASR} & \textbf{AVSR} \\
    \hline
    \multicolumn{7}{c}{\textbf{\textit{Supervised Learning}}} \\
    \hline
    V2P~\cite{shillingford2019large} & \xmark & \xmark & 3,886 & 55.1 & -- & -- \\
    RNN-T~\cite{makino2019recurrent} & \xmark &\cmark & 31,000 & 33.6 & 4.8 & 4.5 \\
    VTP~\cite{prajwal2022sub}  & \xmark & \xmark & 2,676 & 30.7 & -- & -- \\
    Auto-AVSR~\cite{ma2023auto} & \xmark & \xmark & 1,902 & 23.5 & 1.0 & 1.0 \\
    Auto-AVSR~\cite{ma2023auto} & \xmark & \xmark & 3,448 & 19.1 & 1.0 & 0.9 \\
    ViT3D-CM~\cite{serdyuk2022transformer}  & \xmark & \xmark & 90,000 & 17.0 & -- & 1.6 \\
    SynthVSR~\cite{liu2023synthvsr} & \xmark & \xmark & 6,720 & 16.9 & -- & -- \\
    LP Conf~\cite{chang2024conformer} & \xmark & \xmark & 100,000 & 12.8 & -- & 0.9 \\
    \hline
    \multicolumn{7}{c}{\textbf{\textit{Self- or Semi-Supervised Learning}}} \\
    \hline
    {AV-HuBERT$^\dagger$~\cite{shi2022learning}} & {\xmark} & {\xmark} & {1,759} & {26.9} & {--} & {--} \\
    {RAVEn$^\dagger$~\cite{haliassos2022jointly}} & {\xmark} & {\xmark} & {1,759} & {23.1} & {1.4} & {--} \\
    {USR$^\dagger$~\cite{haliassos2024unified}} & {\xmark} & {\cmark} & {1,759} & {22.3} & {1.2} & {1.1} \\

    \hdashline
    AV-HuBERT~\cite{shi2022learning} & \xmark & \xmark & 433 & 28.6 & 1.3 & 1.4 \\
    VATLM~\cite{zhu2023vatlm} & \xmark & \xmark & 433 & 28.4 & -- & 1.2 \\
    RAVEn~\cite{haliassos2022jointly} & \xmark & \xmark & 433 & 28.2 & 1.4 & -- \\
    BRAVEn~\cite{haliassos2024braven} & \xmark & \xmark & 433 & 26.6 & 1.2 & -- \\
    u-HuBERT~\cite{hsu2022u} & \xmark & \cmark & 433 & 29.1 & 1.5 & 1.3 \\
    \hline
    \multicolumn{7}{c}{\textbf{\textit{LLM-Based Models}}} \\
    \hline
    {Llama-AVSR$^\dagger$~\cite{cappellazzo2025large}} & {\xmark} & {\xmark} & {1,759} & {24.0} & {0.79} & {0.77} \\
    {MMS-Llama$^\dagger$~\cite{yeo2025mms}} & {\xmark} & {\xmark} & {1,759} & {--} & {--} & {0.72} \\
    \hdashline
    VSP-LLM~\cite{yeo2024visual} & \xmark & \xmark & 433 & 25.4 & -- & -- \\
    Llama-AVSR~\cite{cappellazzo2025large} & \xmark & \xmark & 433 & 25.3 & 1.1 & 0.95 \\
    MMS-Llama~\cite{yeo2025mms} & \xmark & \xmark & 433 & -- & -- & 0.90 \\
    \hdashline
    Ours & \cmark & \cmark & 433 & 25.9 & 0.79  & 0.93 \\
    \bottomrule
\end{tabular}}
\label{tab:2}
\end{table}

\begin{table}[t]
\centering
\caption{
Ablation study on SLT performance on How2Sign, with progressively added lip encoder and VSR loss.
}
\renewcommand{\arraystretch}{1.3}
\resizebox{0.999\linewidth}{!}{
\begin{tabular}{lccccccc}
\toprule
\multirow{2}{*}{\textbf{Setting}} 
& \multicolumn{2}{c}{\textbf{Encoders}} 
& \multirow{2}{*}{\textbf{VSR Loss}} 
& \multirow{2}{*}{\textbf{BLEU-4}} 
& \multirow{2}{*}{\textbf{ROUGE}} 
& \multirow{2}{*}{\textbf{BLEURT}} \\
& Sign & Lip &  \\
\midrule
1. Base                          & \cmark & \xmark & \xmark & 12.4 & 32.6 & 44.6 \\
2. + Lip Encoder                & \cmark & \cmark & \xmark & 13.9 & 35.0 & 46.2 \\
3. + VSR Loss                   & \cmark & \cmark & \cmark & \textbf{14.4} & \textbf{35.6} & \textbf{46.4} \\
\bottomrule
\end{tabular}
}

\label{tab:3}
\end{table}

\begin{figure}[t]
\centering
\centerline{\includegraphics[width=9cm]{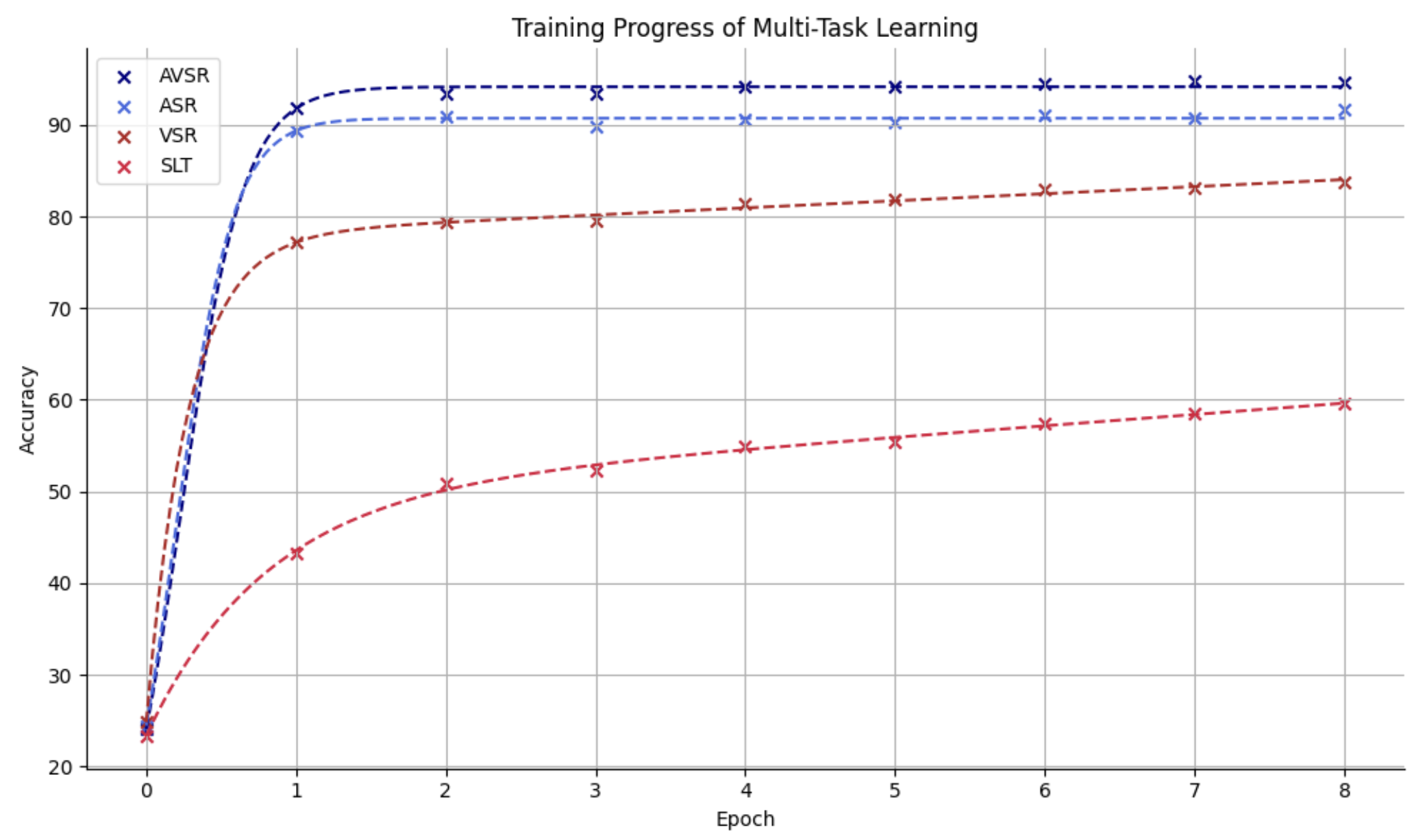}}
\caption{Multi-task training progress for AVSR, ASR, VSR, and SLT. Curves show next-token prediction accuracy per epoch. Audio-related tasks are plotted in blue; visual-related tasks in red.}
\label{fig:2}
\vspace{-15pt}
\end{figure}
\begin{table}[t]
\centering
\caption{Effect of our two-stage multi-task training strategy. Stage 1 trains only on visual tasks (VSR, SLT), while Stage 2 jointly trains all tasks.}
\renewcommand{\tabcolsep}{1mm}
\begin{tabular}{cccccc}
\toprule
\multicolumn{2}{c}{\textbf{Train stage}} & 
\multicolumn{3}{c}{\textbf{WER (\%)}} & 
\multirow{2.5}{*}{\textbf{BLEU-4}} \\
\cmidrule(lr){1-2} \cmidrule(lr){3-5}
\textbf{Stage 1} & \textbf{Stage 2} & \textbf{VSR} & \textbf{ASR} & \textbf{AVSR} & \\
\midrule
\xmark & \cmark & 28.3 & 0.87 & \textbf{0.93} & 14.2 \\
\cmark & \xmark & 25.9 & --   & --   & 14.4 \\
\cmark & \cmark & \textbf{25.9} & \textbf{0.79} & \textbf{0.93} & \textbf{15.2} \\
\bottomrule
\end{tabular}

\label{tab:4}
\end{table}

\begin{table}[t]
\centering
\caption{Effect of modality-specific dropout probabilities on multi-task learning performance.}
\resizebox{0.999\linewidth}{!}{
\begin{tabular}{ccc ccc cccc}
\toprule
\multicolumn{3}{c}{\textbf{Probability}} & \multicolumn{3}{c}{\textbf{WER (\%)}} & 
\multirow{2.5}{*}{\textbf{BLEU-4}} & 
\multirow{2.5}{*}{\textbf{ROUGE}} & 
\multirow{2.5}{*}{\textbf{BLEURT}} \\
\cmidrule(lr){1-3} \cmidrule(lr){4-6}
\textbf{VSR} & \textbf{ASR} & \textbf{AVSR} & \textbf{VSR} & \textbf{ASR} & \textbf{AVSR} & & & \\
\midrule
1/3 & 1/3 & 1/3 & \textbf{25.4} & 1.0 & 1.0 & 15.0 & \textbf{36.4} & 46.8 \\
1/2 & 1/4 & 1/4 & \textbf{25.4} & 0.86 & 1.1 & 15.0 & 36.2 & 46.5 \\
1/4 & 1/2 & 1/4 & 28.2 & 0.89 & 0.93 & 14.7 & 35.6 & 46.3 \\
1/4 & 1/4 & 1/2 & 25.9 & \textbf{0.79} & \textbf{0.93} & \textbf{15.2} & 36.2 & \textbf{47.1} \\
\bottomrule
\end{tabular}
}
\label{tab:5}
\end{table}

\subsection{Ablation study}
\subsubsection{Effectiveness of Lip-Aware SLT via VSR Modeling}
To assess the effectiveness of our lip-aware SLT framework, which incorporates auxiliary VSR modeling within a unified architecture, we conduct controlled ablation experiments by incrementally adding each component. Four configurations are evaluated, as summarized in Table~\ref{tab:3}.

The \textbf{base} configuration includes the sign encoder, sign length adapter, and LLM decoder, trained solely on the SLT task using the How2Sign dataset. This setup yields a BLEU-4 score of 12.4, ROUGE of 32.6, and BLEURT of 44.6. We then add a \textbf{lip encoder} that processes the signer’s mouth region and injects this representation into the LLM alongside sign features. This addition leads to substantial gains: +1.5 BLEU-4, +2.4 ROUGE, and +1.6 BLEURT (13.9 / 35.0 / 46.2), highlighting the importance of explicitly modeling non-manual cues such as lip movements. Finally, we incorporate a \textbf{VSR loss} to train our framework by leveraging an external LRS3 dataset featuring non-signers. This additional supervision further improves the scores to 14.4 BLEU-4, 35.6 ROUGE, and 46.4 BLEURT. We attribute this improvement to the auxiliary task encourages the LLM to internalize the grammar, vocabulary, and contextual semantics of the spoken target language, thereby enhancing its ability to generate well-formed translations from visual sign inputs. This aligns with findings from recent work~\cite{zhang2024scaling}, which showed that jointly training machine translation with SLT improves overall translation performance.

\subsubsection{Effectiveness of Two-stage Multi-Task Learning}
As shown in Figure~\ref{fig:2}, preliminary multi-task training experiments revealed significant convergence imbalances: audio-based tasks (ASR, AVSR) quickly surpassed 90\% next-token prediction accuracy within 2–3 epochs, while visual tasks (VSR, SLT) improved more slowly. Motivated by these observations, we propose a two-stage multi-task learning strategy. To evaluate the effectiveness of this approach, we conducted three configurations: (1) all tasks trained jointly from scratch; (2) only Stage 1 training; and (3) full two-stage training. Results are summarized in Table~\ref{tab:4}.

When trained without Stage 1, the model achieved WERs of 28.3\%, 0.87\%, and 0.93\% for VSR, ASR, and AVSR, respectively, and a BLEU-4 score of 14.2 for SLT. The WER of 28.3\% on the VSR task is noticeably worse than the state-of-the-art result of 25.3\% of Llama-AVSR, suggesting that optimizing all four tasks simultaneously presents challenges. Training with only Stage 1 improved WER to 25.9\% on the VSR task, and SLT BLEU-4 to 14.4, comparing or slightly surpassing task-specific baselines (25.3\% and 13.7, respectively). Finally, the full two-stage training resulted in WERs of 25.9\%, 0.79\%, and 0.93\% for VSR, ASR, and AVSR, and the best SLT BLEU-4 of 15.2. These results demonstrate that the two-stage training approach maintains strong performance in VSR and AVSR while further improving ASR and SLT.

\subsubsection{Impact of Modality Dropout on Multi-Task Learning}
Since the LRS3 dataset provides paired audio-visual inputs, dropping the lip modality corresponds to performing ASR, while dropping the audio modality corresponds to performing VSR. Based on this fact, we leverage modality dropout~\cite{shi2022learning} as an implicit mechanism to control the sampling ratio of each task during training. 

To study its impact, we conduct four experiments using the model pre-trained in Stage 1. Each configuration applies a different ratio for VSR, ASR, and AVSR: (1) uniform (1/3, 1/3, 1/3), (2) VSR-focused (1/2, 1/4, 1/4), (3) ASR-focused (1/4, 1/2, 1/4), and (4) AVSR-focused (1/4, 1/4, 1/2). A notable observation in the ASR-focused setting is that the WER on the VSR task degrades significantly to 28.2\%. Moreover, this configuration yields the lowest SLT performance, with a BLEU-4 score of 14.7. In contrast, configurations with balanced or increased exposure to visual inputs (i.e., VSR/AVSR-focused) consistently yield lower WERs for VSR, in the 25\% range and slightly better SLT performance. While the SLT scores across configurations range from 14.7 to 15.2, indicating some degree of robustness, these results indicate that overemphasis on ASR can negatively affect both recognition and translation quality.

\begin{table}[t]
\centering
\caption{
Performance comparison between task-specific training and joint multi-task training using our proposed framework. Each of the first four rows shows the result when the model is trained on a single task only.
}
\renewcommand{\tabcolsep}{1mm}
\begin{tabular}{cccc|ccc|c}
\toprule
\multicolumn{4}{c|}{\textbf{Task}} & 
\multicolumn{3}{c|}{\textbf{WER (\%)}} & 
\multirow{2.5}{*}{\textbf{BLEU-4}} \\
\cmidrule(lr){1-4} \cmidrule(lr){5-7}
\textbf{SLT} & \textbf{VSR} & \textbf{ASR} & \textbf{AVSR} & \textbf{VSR} & \textbf{ASR} & \textbf{AVSR} & \\
\midrule
\cmark & \xmark & \xmark & \xmark & -- & -- & -- & 12.4 \\
\xmark & \cmark & \xmark & \xmark & \textbf{25.6} & -- & -- & -- \\
\xmark & \xmark & \cmark & \xmark & -- & 1.0 & -- & -- \\
\xmark & \xmark & \xmark & \cmark & -- & -- & \textbf{0.90} & -- \\
\hdashline
\cmark & \cmark & \cmark & \cmark & 25.9 & \textbf{0.79} & 0.93 & \textbf{15.2} \\
\bottomrule
\end{tabular}

\label{tab:6}
\end{table}

\subsubsection{Task-Specific Training Results}
To verify the benefit of multi-task learning in our unified framework, we trained four additional task-specific models: SLT, VSR, ASR, and AVSR, each using only the corresponding task-specific objective. In contrast, the multi-task model is trained using our two-stage multi-task learning strategy. As shown in Table~\ref{tab:6}, the multi-task model achieves substantial improvements in SLT and ASR, with a BLEU-4 score of 15.2 and a WER of 0.79, respectively. While VSR and AVSR show slightly higher WERs than their task-specific counterparts (by 0.3 and 0.03, respectively), the performance remains comparable, demonstrating that our unified model effectively handles all tasks within a single architecture.

\begin{figure*}[t]
\centering
\centerline{\includegraphics[width=18cm]{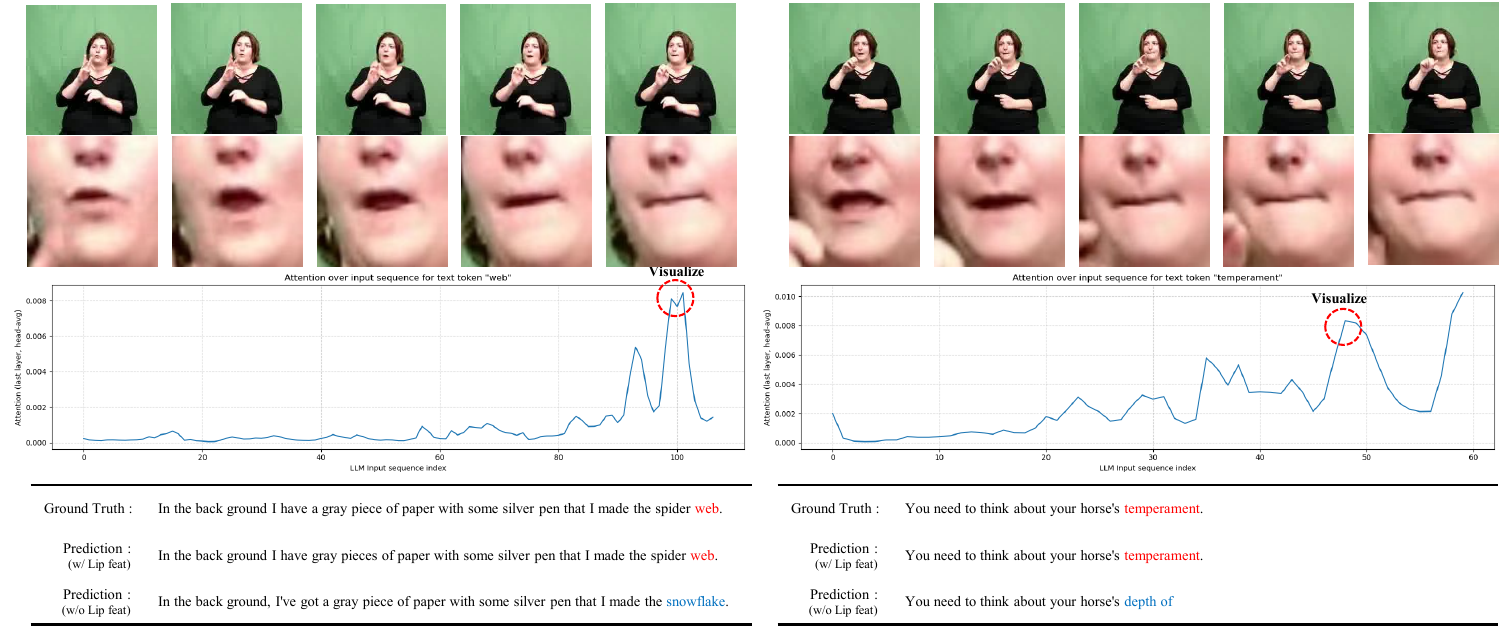}}
\caption{Visualization of attention distributions and qualitative comparisons of SLT predictions with and without lip features. From top to bottom: (1) signer frames, (2) corresponding lip-region crops, (3) attention scores averaged over all heads in the last LLM layer, and (4) ground-truth and predicted sentences. Examples on the left and right correspond to the text tokens “web” and “temperament,” respectively. }
\label{fig:3}
\end{figure*}

\subsubsection{Qualitative Results}
To visually verify the effectiveness of lip-movement modeling in the SLT task, we analyze the attention scores averaged over all heads in the last layer with respect to the LLM input sequence index (x-axis), as shown in Figure~\ref{fig:3}. We also visualize the input frames corresponding to regions with high attention scores, providing an intuitive comparison between the model’s focus and the visual cues in the sign and lip movements.

On the left side of Figure~\ref{fig:3}, qualitative examples compare predictions with and without lip features. In the example, the ground-truth sentence is “In the background I have gray pieces of paper with some silver pen that I made the spider web.” Without lip features, the model incorrectly predicts “...that I made the snowflake,” failing to capture the intended meaning. When lip features are included, it correctly outputs “web”. The corresponding high-attention regions align with mouth movements resembling the articulation of “web”. On the right side, another example shows the ground-truth sentence “You need to think about your horse’s temperament.” Without lip features, the model produces the incorrect phrase “...your horse’s depth of,” whereas with lip features it accurately generates “temperament.” These results visually confirm that visual cues from lip movements provide critical articulatory evidence for distinguishing semantically similar expressions.

\begin{figure}[t]
\centering
\centerline{\includegraphics[width=9cm]{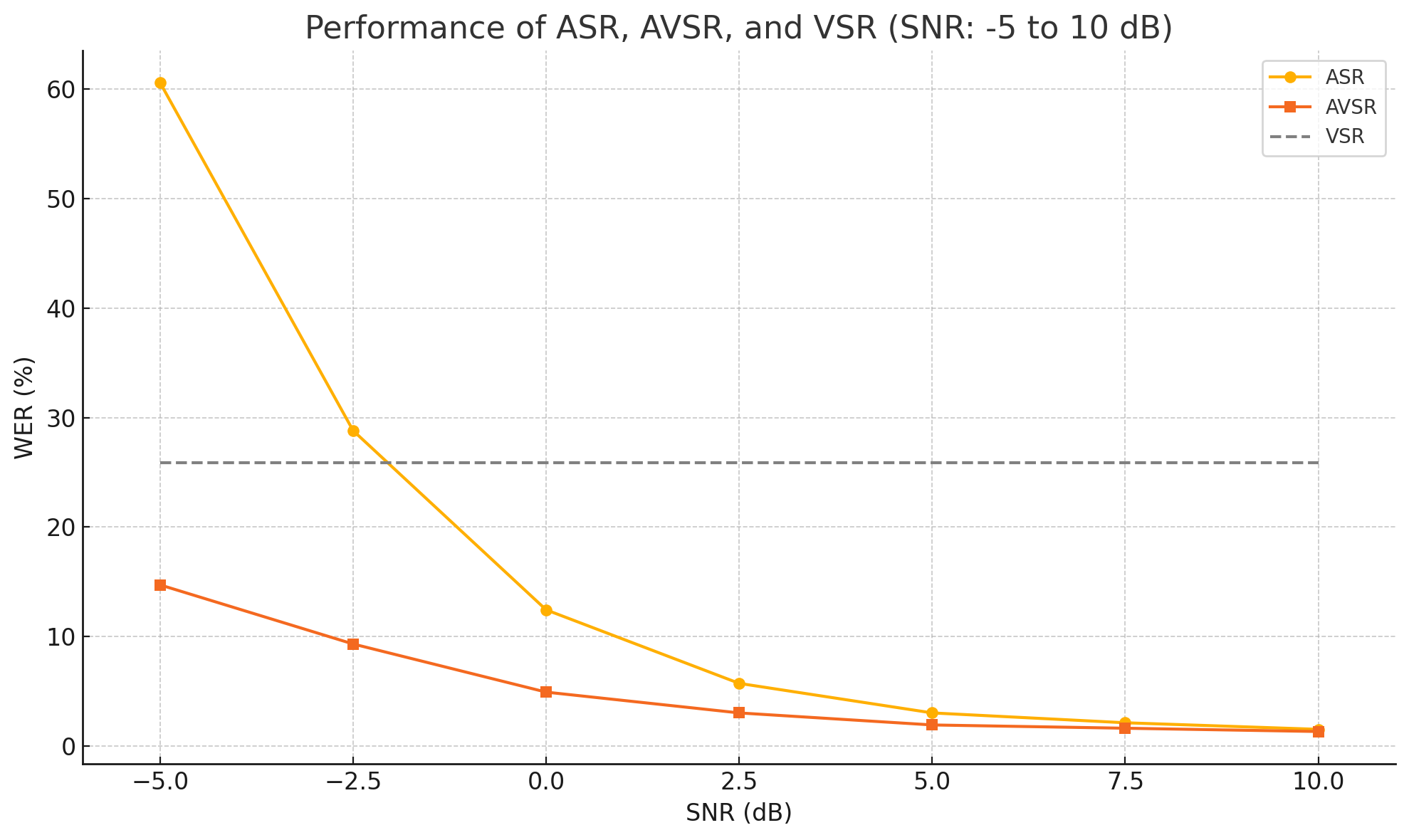}}
\caption{WER comparison of ASR, AVSR, and VSR under varying SNR levels (-5 to 10 dB) using babble noise on the LRS3 dataset. All results are obtained from a single unified model evaluated across the three tasks.}
\label{fig:4}
\end{figure}
\subsubsection{Noise Robustness Evaluation }
To examine the noise robustness of our unified framework, we evaluate ASR and AVSR performance under varying levels of additive babble noise, ranging from -5 dB to 10 dB SNR in 2.5 dB increments. As shown in Figure~\ref{fig:4}, the AVSR model leverages the complementary nature of audio and lip movements to maintain high accuracy even in challenging acoustic conditions. While ASR and AVSR achieve similar performance in the high-SNR regime (5-10 dB), the performance gap widens as SNR decreases. For instance, at 2.5 dB, AVSR achieves 3.0\% WER compared to 5.7\% for ASR; at 0 dB, the gap further increases (4.9\% vs 12.4\%), and at -2.5 dB, AVSR maintains 9.3\% WER while ASR degrades to 28.8\%. These results demonstrate the robustness of our model in noisy environments. Additionally, AVSR consistently outperforms VSR across all noise levels, indicating that even degraded audio provides complementary information beyond what visual input alone can offer.

\section{Conclusion}
We introduced a unified tri-modal framework that jointly models sign language, lip movements, and audio for spoken-language text generation across SLT, VSR, ASR, and AVSR tasks. By designing a modality-aligned architecture with a shared linguistic representation and a multi-task training strategy, our system achieved competitive performance across all tasks, matching or surpassing task-specific state-of-the-art models. Furthermore, our analysis demonstrated that explicitly modeling lip movements as a distinct modality significantly improved SLT performance, validating their role as essential non-manual cues in sign language understanding.


\section*{Acknowledgments}
Research supported by the NVIDIA Academic Grant Program using NVIDIA A100 Tensor Core GPUs, CUDA Toolkit, and NCCL.

\bibliographystyle{IEEEtran}
\bibliography{main}
\newpage
\begin{IEEEbiography}[{\includegraphics[width=1in,height=1.25in,clip]{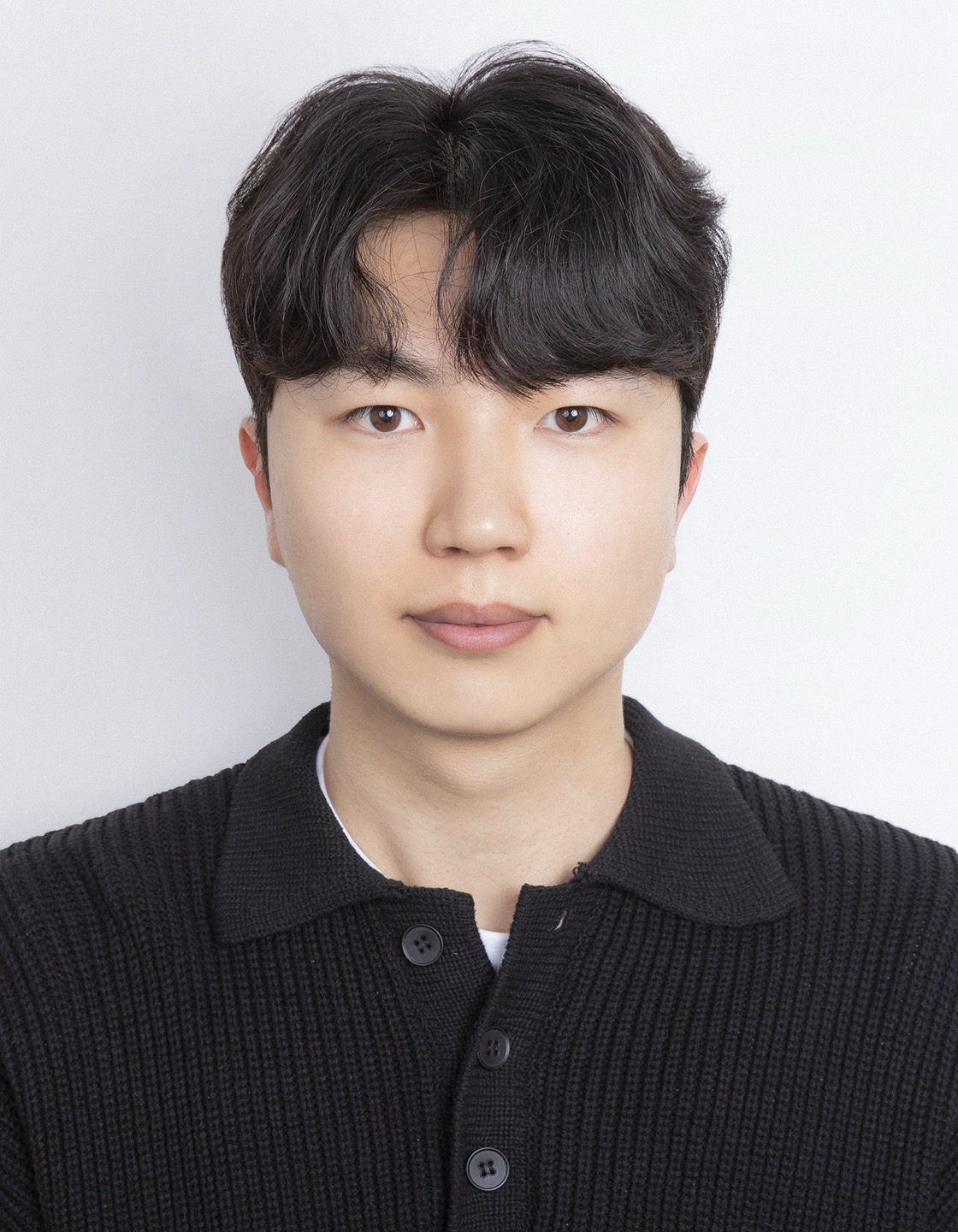}}]{Jeong Hun Yeo}
received the B.S. and M.S. degrees in electrical \& electronic engineering from Korea Advanced Institute of Science and Technology (KAIST), Daejeon, South Korea, in 2020 and 2022, respectively. He is currently pursuing the Ph.D. degree in electrical engineering at KAIST, Daejeon, South Korea. His research interests include deep learning, image/video analysis, visual speech recognition, and multi-modal learning.
\end{IEEEbiography}
\vspace{-1cm}

\begin{IEEEbiography}[{\includegraphics[width=1in,height=1.25in,clip]{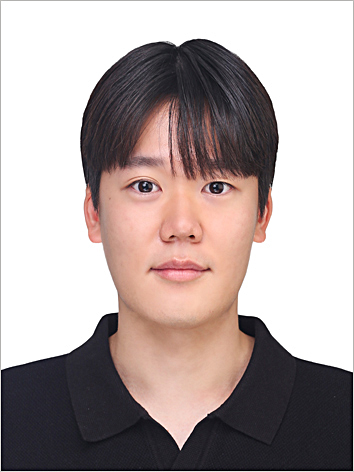}}]{Hyeongseop Rha}
received the BS degree in electrical \& electronic engineering from Yonsei University, Seoul, South Korea. He is currently pursuing the integrated MS/PhD degree in the School of Electrical Engineering at the Korea Advanced Institute of Science and Technology (KAIST), Daejeon, South Korea. His research interests include multi-modal learning, multi-modal large language models, and multi-modal human interaction.
\end{IEEEbiography}
\vspace{-1cm}

\begin{IEEEbiography}[{\includegraphics[width=1in,height=1.25in,clip]{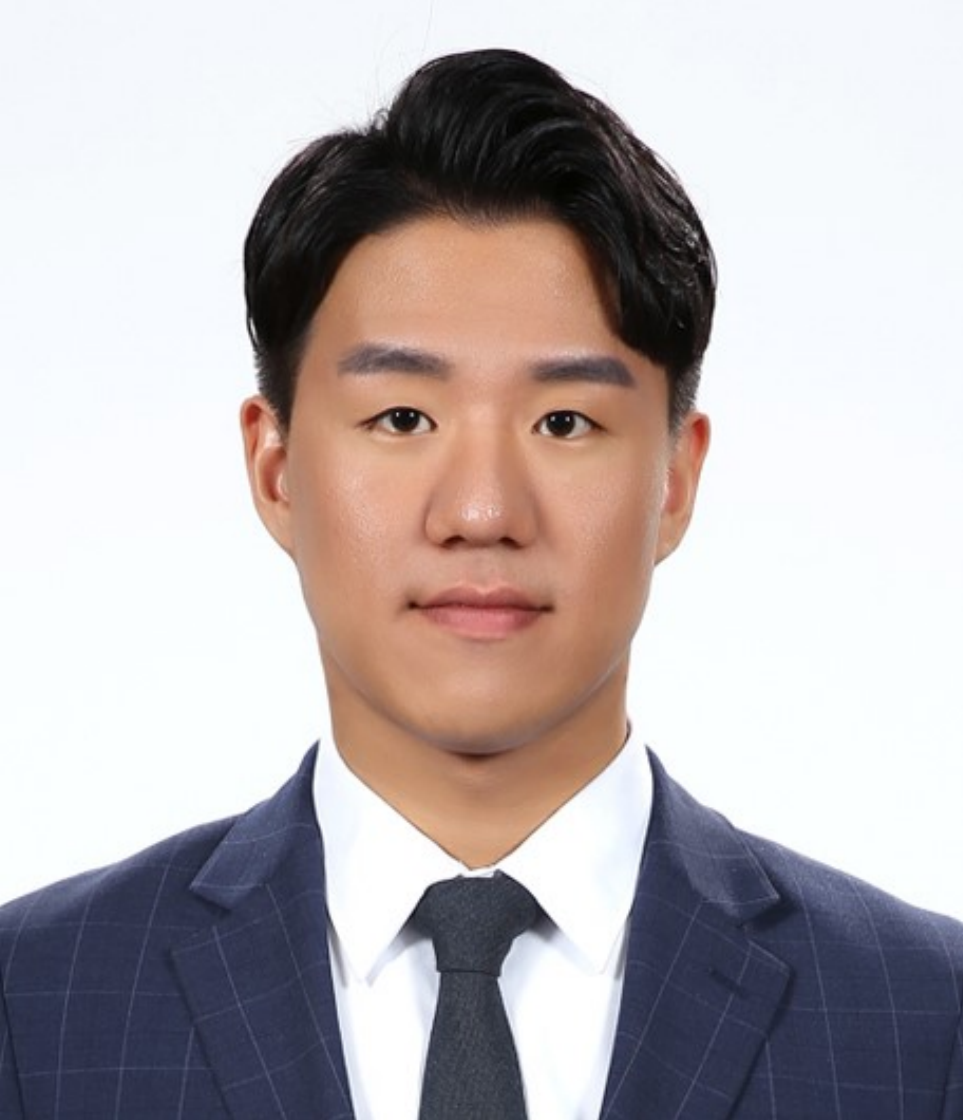}}]{Sungjune Park}
received the B.S. degree in school of electrical and electronic engineering from Yonsei University, Seoul, South Korea, in 2019. He is currently working toward his Ph.D. degree in electrical engineering, Korea Advanced Institute of Science and Technology (KAIST), Daejeon, South Korea. His research interests include machine learning, deep learning, image-text retrieval, adversarial robustness, and object detection. These days, he is currently interested in taking advantage of language into computer vision AI models and developing robust multimodal large language models in wild real-world environments.
\end{IEEEbiography}
\vspace{-1cm}

\begin{IEEEbiography}[{\includegraphics[width=1in,height=1.25in,clip]{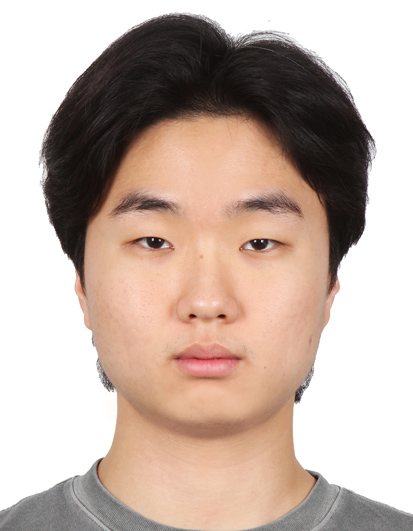}}]{Junil Won}
received his B.S. degree in Electrical Engineering from the Korea Advanced Institute of Science and Technology (KAIST), Daejeon, South Korea. His research interests include deep learning,  human-centric understanding in Multimodal Large Language Models(MLLMs), human-AI interaction, and agentic AI.
\end{IEEEbiography}
\vspace{-1cm}

\begin{IEEEbiography}[{\includegraphics[width=1in,height=1.25in,clip]{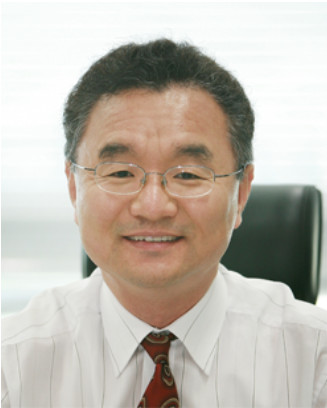}}]{Yong Man Ro}
(Senior Member, IEEE) received a B.S. degree from Yonsei University, Seoul, South Korea, and a M.S. and Ph.D. degrees from the Korea Advanced Institute of Science and Technology (KAIST), Daejeon, South Korea. He was a Researcher at Columbia University, a Visiting Researcher at the University of California at Irvine, Irvine, CA, USA, and a Research Fellow of the University of California at Berkeley, Berkeley, CA, USA. He was a Visiting Professor with the Department of Electrical and Computer Engineering, University of Toronto, Canada. He is currently a Professor at the Department of Electrical Engineering and the Director of the Center for Applied Research in Artificial Intelligence (CARAI), KAIST. Among the years, he has been conducting research in a wide spectrum of image and video systems research topics. Among those topics, his interests include image processing, computer vision, visual recognition, multimodal learning, video representation/compression, and object detection. He received the Young Investigator Finalist Award of ISMRM, in 1992, and the Year’s Scientist Award (Korea), in 2003. He served as an Associate Editor for IEEE SIGNAL PROCESSING LETTERS. He was an Associate Editor for IEEE TRANSACTIONS ON CIRCUITS AND SYSTEMS FOR VIDEO TECHNOLOGY and currently serves as an Associate Editor for IEEE TRANSACTIONS ON IMAGE PROCESSING. He served as a TPC in many international conferences, including the program chair, and organized special sessions.
\end{IEEEbiography}

\vfill

\end{document}